\documentclass{article}
\usepackage{spconf,amsmath,graphicx,hyperref,afterpage,booktabs,array,multirow,amssymb}
\usepackage[utf8]{inputenc}

\title{Enhancing Cross-View Geo-Localization Generalization via Global-Local Consistency and Geometric Equivariance}
\name{
\begin{tabular}{c}
Xiaowei Wang$^{1}$, 
Di Wang$^{1}$\sthanks{Corresponding author.}, 
Ke Li$^{1}$, 
Yifeng Wang$^{1}$ 
Chengjian Wang$^{2}$, 
Libin Sun$^{1}$, \\
Zhihong Wu$^{1}$, 
Yiming Zhang$^{3}$, 
Quan Wang$^{1}$
\end{tabular}
}

\address{
    $^{1}$ School of Computer Science and Technology, Xidian University, 710126, China.\\
    $^{2}$ School of Artificial Intelligence, Xidian University, 710126, China.\\
    $^{3}$ Department of Mathematics, University of California San Diego, USA.
}
\begin{document}
%
\maketitle
\begin{abstract}
Cross-view geo-localization (CVGL) aims to match images of the same location captured from drastically different viewpoints. Despite recent progress, existing methods still face two key challenges: (1) achieving robustness under severe appearance variations induced by diverse UAV orientations and fields of view, which hinders cross-domain generalization, and (2) establishing reliable correspondences that capture both global scene-level semantics and fine-grained local details.
In this paper, we propose EGS, a novel CVGL framework designed to enhance cross-domain generalization. Specifically, we introduce an E(2)-Steerable CNN encoder to extract stable and reliable features under rotation and viewpoint shifts. Furthermore, we construct a graph with a virtual super-node that connects to all local nodes, enabling global semantics to be aggregated and redistributed to local regions, thereby enforcing global–local consistency.
Extensive experiments on the University-1652 and SUES-200 benchmarks demonstrate that EGS consistently achieves substantial performance gains and establishes a new state of the art in cross-domain CVGL.
\end{abstract}

\section{Introduction}
\label{sec:intro}

Cross-View Geo-Localization (CVGL) aims to determine geographic locations by leveraging semantic correspondences between UAV and satellite imagery. It has emerged as a critical capability for applications such as autonomous navigation, GPS correction, and emergency search, particularly in scenarios where GNSS signals are weak or unavailable. Recent research commonly formulates CVGL as a cross-view image retrieval task, in which a UAV query image is matched against a large satellite database with known geographic coordinates to achieve accurate localization.

The core challenges of CVGL lie in extreme viewpoint variations leading to geometric mismatches, discrepancies in imaging time and resolution, and real-world factors such as orientation uncertainty and field-of-view (FoV) cropping. A particularly critical issue is performance degradation under cross-area or cross-dataset settings, where models often overfit low-level details or dataset biases and thus fail to maintain rotational invariance and feature consistency. To address these limitations, recent studies have pursued different directions. Some focus on representation learning, such as ConGeo \cite{congeo} disrupting north-alignment shortcuts through contrastive learning, GeoDTR \cite{geodtr, geodtr+} disentangling geometric layout from appearance and augmenting with layout simulation, and CVD \cite{cvd} separating content and viewpoint to alleviate viewpoint conflicts. Others advance feature modeling, where Transformer-based methods \cite{transgeo} emphasize global semantics with attention and positional encoding, while local or patch-wise methods \cite{fsra, lpn, sdpl} enhance discriminativeness by mining object-centric relations. In addition, training strategies such as symmetric InfoNCE and hard negative sampling \cite{sample4geo} further improve generalization without geometric preprocessing. MCCG \cite{mccg} strengthens cross-dimensional interactions for richer representations, while new datasets and protocols \cite{game4loc, university1652, sues200, vigor, cvusa, cvact} introduce more realistic evaluation settings, pushing CVGL closer to practical deployment. Multimodal retrieval studies \cite{mdbe}, highlight the value of discriminability, yet such aspects remain underexplored in CVGL. Despite these advances, progress remains limited, as many methods still overlook fundamental issues that hinder effective cross-domain generalization.


\begin{figure}
    \centering
    \includegraphics[width=1\linewidth]{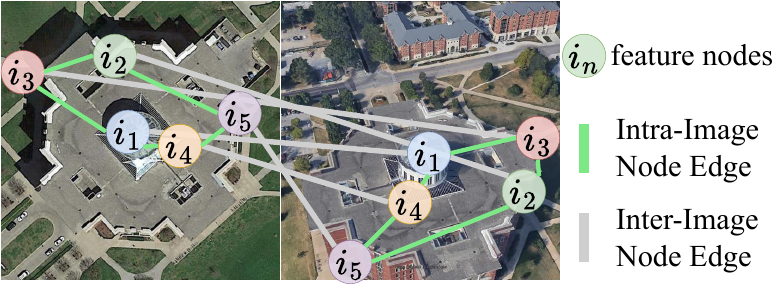}
    \caption{Graph-based CVGL representation, where feature nodes encode image regions, intra-image edges capture local structure, and inter-image edges model cross-view correspondences.}
    \label{fig:fig1}
\end{figure}

Nevertheless, existing mainstream frameworks still suffer from two common limitations:
\textit{(1) Sensitivity to orientation and rotation.} Even with random rotation and FoV perturbations during training, invariance or equivariance to geometric transformations is often acquired passively, making models unstable under unknown orientations and scale variations, a limitation especially pronounced in UAV flight applications.
\textit{(2) Lack of explicit global-local consistency modeling.} Many approaches rely on global pooling or rule-based patch partitioning. These strategies may fail under domain/area shifts due to local detail drift, or they disrupt object continuity during partitioning, limiting joint expression of global semantics and local details.

To overcome the \textit{limitation (1)}, we explicitly embed rotational consistency into the feature encoding stage by constraining it with equivariant priors to Euclidean transformations \cite{e2steerable}. Under this design, variations in viewpoint or field of view induce predictable transformations in the latent space, ensuring stable performance across arbitrary orientations. Unlike invariance acquired implicitly through data augmentation, this approach provides a structural inductive bias that yields more reliable cross-view correspondences.

To address the \textit{limitation (2)}, we reformulate CVGL representations as graphs, where image regions are represented as nodes and semantic or spatial relations define the edges. An illustration of this graph-based formulation, including intra-image and inter-image node relationships, is shown in Fig.\ref{fig:fig1}. In addition, a virtual super node is introduced, initialized via global pooling and bidirectionally connected to all nodes. Through iterative message passing, this super node aggregates global context and redistributes it to local regions, thereby structurally enforcing global–local consistency and enhancing generalization across domain shifts.

Based on these two designs, we present \textbf{EGS}, a unified framework for CVGL.  By jointly enforcing global-local consistency and rotation-equivariant encoding, EGS achieves stable and transferable representations across challenging UAV-satellite scenarios. To summarise our contribution in this work:
\begin{itemize}
\item We propose a novel unified framework, \textbf{EGS}, for CVGL that enhances generalization in challenging conditions.
\item We introduce two core modules, including a graph-based super node mechanism to enforce global-local consistency and an equivariant encoding design to model rotational consistency.
\item Extensive experiments validate the effectiveness of the proposed \textbf{EGS} across domain scenarios.
\end{itemize}

\begin{figure*}
    \centering
    \includegraphics[width=1\linewidth]{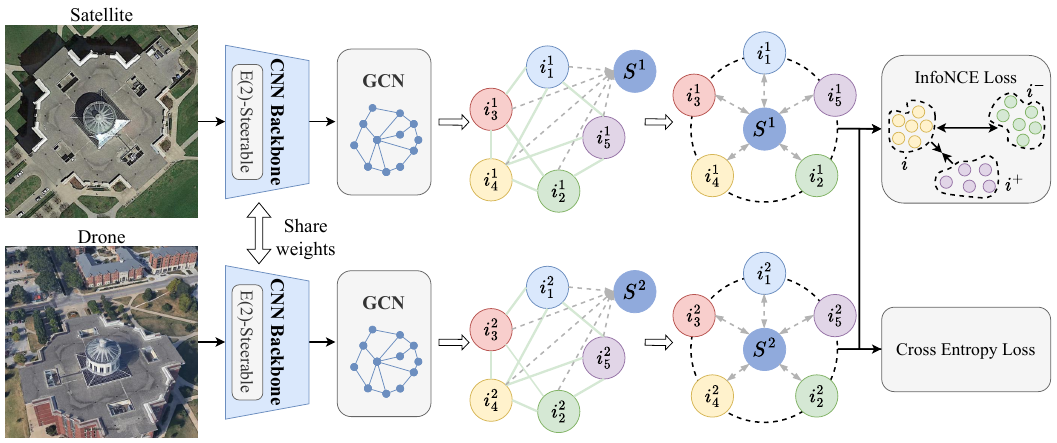}
    \caption{Overview of the proposed \textbf{EGS} framework for UAV-satellite cross-view geo-localization. }
    \label{fig:fig2}
\end{figure*}

\section{Method}
\label{sec:pagestyle}

\subsection{Overall Framework}

We propose \textbf{EGS}, a unified framework for UAV-satellite CVGL that integrates rotation-equivariant feature encoding with graph-based representations and a virtual super node mechanism. As illustrated in Fig.  \ref{fig:fig2}, given a satellite–drone image pair, features are extracted by a shared CNN backbone with equivariant design, then transformed into graph structures to model local patch relations and global semantics. A virtual super node aggregates and redistributes contextual information, and the final representations are optimized using contrastive and classification objectives to achieve reliable alignment across views.

Given a satellite or drone image, we first employ a shared CNN backbone to extract feature maps. These maps are partitioned into $N$ patches, each treated as a node encoding semantic or structural cues. The image is thus represented as a graph $G=(V,E)$, where
\begin{equation}
\label{equ1}
V = \{x_1, x_2, \dots, x_N\}, \quad 
A_{ij} = 
\begin{cases} 
1, & \text{if } i \text{ is adjacent to } j \\
0, & \text{otherwise}.
\end{cases}
\end{equation}
This formulation captures intra-image dependencies beyond the regular CNN grid.

\subsection{Rotation-Equivariant Feature Extraction}

CVGL often involves large orientation and field-of-view differences between UAV and satellite imagery. To address this, we incorporate a rotation-equivariant design into the backbone. Given a rotation matrix $R$, the steerable convolution ensures predictable feature transformation:

\begin{equation}
    \label{equ2}
\hat{f}(\mathbf{x}, R) = \sum_{k} \phi_k(\mathbf{x}) \cdot W_k(R),    
\end{equation}
where $\phi_k(\mathbf{x})$ is the input feature at location $\mathbf{x}$, and $W_k(R)$ denotes rotation-equivariant filters. This constraint introduces an architectural inductive bias, making features stable under arbitrary orientations.

\subsection{Graph-Based Representation with Super Node}

To enforce global-local consistency, we introduce a virtual super node. It is initialized via global average pooling of all patch features:

\begin{equation}
    \label{equ3}
\mathbf{s_0} = \frac{1}{N} \sum_{i=1}^{N} \mathbf{x_i}.
\end{equation}
The super node is bidirectionally connected to all nodes, leading to an augmented adjacency matrix $\tilde{A}$:

\begin{equation}
    \label{equ4}
\tilde{A}_{ij} = 
\begin{cases} 
1, & \text{if nodes } i \text{ and } j \text{ are connected } \\
0, & \text{otherwise}.
\end{cases}
\end{equation}
During message passing, each node updates its embedding as:
\begin{equation}
    \label{equ5}
h_i^{(l+1)} = \sigma \left( \sum_{j \in \mathcal{N}(i)} \frac{A_{ij}}{d_j} h_j^{(l)} W^{(l)} \right),
\end{equation}
where $d_j$ is the degree of node $j$ and $W^{(l)}$ is the trainable weight. After multiple propagation steps, both local nodes and the super node encode joint global-local semantics.

Finally, the global descriptor is obtained either from the super node embedding $\mathbf{s_L}$ or by concatenating it with refined patch features:
\begin{equation}
    \label{equ6}
\mathbf{z} = \mathbf{s_L} \quad \text{or} \quad \mathbf{z} = [h_1^{(L)}, h_2^{(L)}, \dots, h_N^{(L)}, \mathbf{s_L}].
\end{equation}

\subsection{Training Objective}

We jointly optimize the model using contrastive learning and cross-entropy supervision. The contrastive loss (InfoNCE) encourages satellite-drone pairs from the same location to have similar representations while pushing apart mismatched pairs. The cross-entropy loss supervises classification consistency based on the global descriptor.
The total loss is:

\begin{equation}
    \label{equ7}
\mathcal{L} = \mathcal{L}_{\text{InfoNCE}} + \mathcal{L}_{\text{CE}}.
\end{equation}
This dual-objective training enhances both discriminative ability and cross-view alignment.


\section{Experiments}
\label{sec:typestyle}

\begin{table*}[ht]
\centering
\resizebox{\linewidth}{!}{
\begin{tabular}{l*{16}{c}}
\toprule
& \multicolumn{8}{c}{\textbf{Drone $\rightarrow$ Satellite}} & \multicolumn{8}{c}{\textbf{Satellite $\rightarrow$ Drone}} \\
\cmidrule(lr){2-9} \cmidrule(l){10-17}
\textbf{Method} & \multicolumn{2}{c}{\textbf{150m}} & \multicolumn{2}{c}{\textbf{200m}} & \multicolumn{2}{c}{\textbf{250m}} & \multicolumn{2}{c}{\textbf{300m}} & \multicolumn{2}{c}{\textbf{150m}} & \multicolumn{2}{c}{\textbf{200m}} & \multicolumn{2}{c}{\textbf{250m}} & \multicolumn{2}{c}{\textbf{300m}} \\
\cmidrule(lr){2-3} \cmidrule(lr){4-5} \cmidrule(lr){6-7} \cmidrule(lr){8-9} \cmidrule(lr){10-11} \cmidrule(lr){12-13} \cmidrule(lr){14-15} \cmidrule(lr){16-17}
\textbf{ } & \textbf{AP} & \textbf{R@1} & \textbf{AP} & \textbf{R@1} & \textbf{AP} & \textbf{R@1} & \textbf{AP} & \textbf{R@1} & \textbf{AP} & \textbf{R@1} & \textbf{AP} & \textbf{R@1} & \textbf{AP} & \textbf{R@1} & \textbf{AP} & \textbf{R@1} \\
\midrule
University & 35.83 & 29.98 & 41.32 & 35.45 & 42.02 & 35.95 & 41.38 & 35.20 & 27.38 & 33.75 & 33.18 & 35.00 & 37.00 & 41.25 & 38.02 & 36.25 \\
LPN & 42.83 & 36.70 & 52.99 & 46.72 & 59.42 & 53.62 & 62.15 & 56.55 & 25.30 & 30.00 & 34.36 & 38.75 & 38.53 & 42.50 & 43.92 & 53.75 \\
SDPL & 29.31 & 23.33 & 36.47 & 29.93 & 40.56 & 34.00 & 44.16 & 37.26 & 22.10 & 21.25 & 30.97 & 30.00 & 34.19 & 33.75 & 37.26 & 40.00 \\
\textbf{Ours} & \textbf{47.28} & \textbf{41.83} & \textbf{55.40} & \textbf{49.55} & \textbf{61.35} & \textbf{55.45} & \textbf{63.64} & \textbf{57.53} & \textbf{34.04} & \textbf{41.25} & \textbf{42.71} & \textbf{46.25} & \textbf{47.48} & \textbf{57.50} & \textbf{51.96} & \textbf{62.50} \\
\midrule
Baseline & 35.83 & 29.98 & 41.32 & 35.45 & 42.02 & 35.95 & 41.38 & 35.20 & 27.38 & 33.75 & 33.18 & 35.00 & 37.00 & 41.25 & 38.02 & 36.25 \\
+ GAT & 42.79 & 36.72 & 53.68 & 47.43 & 58.16 & 52.03 & 61.25 & 55.23 & 27.08 & 32.50 & 36.25 & 41.25 & 44.12 & 50.00 & 50.39 & 57.50 \\
+ GCN & 42.80 & 37.28 & 53.94 & 48.55 & 60.78 & 55.08 & 67.50 & 62.92 & 24.79 & 26.25 & 33.93 & 38.75 & 39.72 & 47.50 & 44.56 & 53.75 \\
+ EGS & 47.28 & 41.83 & 55.40 & 49.55 & 61.35 & 55.45 & 63.64 & 57.53 & 34.04 & 41.25 & 42.71 & 46.25 & 47.48 & 57.50 & 51.96 & 62.50 \\

\bottomrule
\end{tabular}}
\caption{Cross-dataset generalization performance, trained on University-1652 and evaluated on SUES-200.}
\label{tab:comparison1}
\end{table*}

\begin{table}[ht]
\centering
\resizebox{0.95\linewidth}{!}{
\begin{tabular}{lccccc}
\toprule
\textbf{Method} & \textbf{AP} & \textbf{R@1} & \textbf{R@5} & \textbf{R@10} & \textbf{R@1\%} \\
\midrule
\textbf{Drone $\rightarrow$ Satellite} \\
University & 21.53 & 17.59 & 32.46 & 40.23 & 41.37 \\
LPN        & 43.01 & 38.90 & 55.39 & 62.97 & 64.10 \\
SDPL       & 46.45 & 42.39 & 59.21 & 66.04 & 67.01 \\
\textbf{Ours} & \textbf{47.33} & \textbf{43.11} & \textbf{60.91} & \textbf{67.85} & \textbf{69.04} \\
\midrule
\textbf{Satellite $\rightarrow$ Drone} \\
University & 16.51 & 31.67 & 42.65 & 48.50 & 81.46 \\
LPN        & 46.12 & 69.76 & 77.32 & 79.03 & 95.72 \\
SDPL       & 44.38 & 59.20 & 65.05 & 67.76 & 87.30 \\
\textbf{Ours} & \textbf{48.48} & \textbf{72.19} & \textbf{79.61} & \textbf{82.17} & \textbf{97.44} \\
\bottomrule
\end{tabular}}
\caption{Cross-dataset generalization performance, trained on SUES-200 and evaluated on University-1652.}
\label{tab:comparison2}
\end{table}

\subsection{Settings}
\noindent \textbf{Datasets}. We evaluate our method on two representative CVGL benchmarks: University-1652 \cite{university1652} and SUES-200 \cite{sues200}. University-1652 includes 1,652 buildings from 72 universities, with images from three platforms-drone, satellite, and ground-designed for tasks such as drone-based localization and navigation. The dataset features synthetic drone images and offers a multi-source, multi-view framework for geo-localization. SUES-200 consists of 200 target scenes, with drone images captured at four different flight heights (150m, 200m, 250m, 300m) and corresponding satellite images, enabling the evaluation of cross-view image matching models at varying altitudes.

\noindent \textbf{Evaluation metrics.} We adopt five standard retrieval metrics to evaluate the performance of cross-view matching: Average Precision (AP), Recall@K (for K=1, 5, 10), and Recall@1\%. These metrics are commonly used to assess the accuracy and effectiveness of image retrieval systems by measuring the relevance of the retrieved results.

\noindent \textbf{Implementation details.} We adopt ResNet-50 \cite{resnet50}, initialized with ImageNet pre-trained weights, as the backbone network for visual feature extraction. All images are resized to 256×256 pixels, with random cropping and horizontal flipping applied during training. The model is trained with SGD (momentum 0.9, weight decay 0.0005, batch size 32), where the initial learning rate is 0.001 for backbone layers and 0.01 for newly added layers. At test time, similarity between queries and gallery images is measured by Euclidean distance. All experiments are conducted in PyTorch on an NVIDIA RTX 4090 GPU with 24GB memory.

\subsection{Cross-Dataset Evaluation}
To evaluate the generalization ability of our model in unseen environments, we conduct cross-dataset evaluations between University-1652 and SUES-200. Table~\ref{tab:comparison1} reports the results when models are trained on University-1652 and tested on SUES-200. Our method consistently surpasses baseline approaches across all flight altitudes in both Drone $\rightarrow$ Satellite and Satellite $\rightarrow$ Drone tasks. For instance, at 300m altitude, it achieves 57.53\% Recall@1 and 63.64\% AP in Drone $\rightarrow$ Satellite retrieval, as well as 62.50\% Recall@1 and 51.96\% AP in Satellite $\rightarrow$ Drone retrieval, demonstrating clear advantages over existing methods.Table~\ref{tab:comparison2} further presents the reverse evaluation, where models are trained on SUES-200 and tested on University-1652. Similar improvements are observed: our approach achieves 43.11\% Recall@1 and 69.04\% Recall@Top1\% in Drone $\rightarrow$ Satellite retrieval, and 72.19\% Recall@1 and 97.44\% Recall@Top1\% in Satellite $\rightarrow$ Drone retrieval, substantially outperforming prior methods. All methods adopt the same ResNet-50 backbone to ensure fairness. These results collectively indicate that our model learns transferable representations that generalize effectively across datasets, yielding significant performance gains in both directions of cross-dataset evaluation.

\subsection{Ablation Studies}
To examine the contribution of individual components in our framework, we perform an ablation study on the cross-dataset setting, where models are trained on University-1652 and tested on SUES-200. As shown in Table~\ref{tab:comparison1}, the baseline model achieves limited performance, while incorporating graph-based modules yields consistent improvements. Specifically, adding GAT or GCN enhances both AP and Recall@1 across all flight altitudes in both Drone $\rightarrow$ Satellite and Satellite $\rightarrow$ Drone tasks, demonstrating the benefit of modeling structural relationships. Furthermore, introducing our EGS framework brings the most significant gains, reaching 63.64\% AP and 57.53\% Recall@1 at 300m in the Drone $\rightarrow$ Satellite scenario, and 51.96\% AP and 62.50\% Recall@1 at 300m in the Satellite $\rightarrow$ Drone scenario. These results clearly validate the effectiveness of each component and highlight the superior contribution of EGS in improving cross-dataset generalization.

\section{Conclusions}
\label{sec:majhead}
In this paper we introduce EGS, a unified framework for CVGL that addresses the challenges of global-local feature consistency and orientation sensitivity. By reformulating cross-view representations into graph structures with a Virtual Super Node for global-local integration, the framework enhances generalization across datasets and maintains consistency under large viewpoint changes. In addition, the use of E(2)-Steerable CNNs ensures stability under unknown orientations and field-of-view variations commonly encountered in UAV scenarios. Extensive experiments on University-1652 and SUES-200 benchmarks validate the effectiveness of our method, demonstrating competitive in-domain performance and significant improvements in cross-domain generalization. Overall, the proposed framework provides a scalable solution for CVGL and strengthens its applicability in real-world UAV navigation and multi-view localization tasks.

\bibliographystyle{IEEEbib}
\bibliography{strings}

\end{document}